\documentclass[conference]{IEEEtran}
\IEEEoverridecommandlockouts
\usepackage{diagbox}
\usepackage{caption}
\usepackage{tabu}
\usepackage{multirow}
\usepackage[table]{xcolor}
\usepackage{tabularx}  
\usepackage{ragged2e}
\usepackage{multicol}
\usepackage{url}
\usepackage{amsmath,amssymb,amsfonts}
\usepackage{graphicx}
\IEEEaftertitletext{\vspace{-1.5\baselineskip}}

\def\BibTeX{{\rm B\kern-.05em{\sc i\kern-.025em b}\kern-.08em
    T\kern-.1667em\lower.7ex\hbox{E}\kern-.125emX}}
\begin{document}

\title{Improving Accented Speech Recognition using Data Augmentation based on Unsupervised Text-to-Speech Synthesis\\
%\title{Unsupervised Text-to-Speech Synthesis for Data Augmentation in Accented Speech Recognition\\
\vspace{-5pt}}
%\vspace{-5pt}}

\author{\IEEEauthorblockN{Cong-Thanh Do}
\IEEEauthorblockA{\textit{Toshiba Research Europe} \\
%\textit{Toshiba Europe Ltd.}\\
Cambridge, UK \\
\footnotesize \texttt{cong-thanh.do@toshiba.eu}}
\and
\IEEEauthorblockN{Shuhei Imai}
\IEEEauthorblockA{\textit{Tohoku University} \\
%\textit{Tohoku University}\\
Sendai, Japan \\
\footnotesize \texttt{shuhei.imai@tohoku.ac.jp}}
\and
\IEEEauthorblockN{Rama Doddipatla}
\IEEEauthorblockA{\textit{Toshiba Research Europe} \\
%\textit{Toshiba Europe Ltd.}\\
Cambridge, UK \\
\footnotesize \texttt{rama.doddipatla@toshiba.eu}}
\and
\IEEEauthorblockN{Thomas Hain}
\IEEEauthorblockA{\textit{University of Sheffield} \\
%\textit{name of organization (of Aff.)}\\
Sheffield, UK \\
\footnotesize \texttt{t.hain@sheffield.ac.uk}}

%\and
%\IEEEauthorblockN{5\textsuperscript{th} Given Name Surname}
%\IEEEauthorblockA{\textit{dept. name of organization (of Aff.)} \\
%\textit{name of organization (of Aff.)}\\
%City, Country \\
%email address or ORCID}
%\and
%\IEEEauthorblockN{6\textsuperscript{th} Given Name Surname}
%\IEEEauthorblockA{\textit{dept. name of organization (of Aff.)} \\
%\textit{name of organization (of Aff.)}\\
%City, Country \\
%email address or ORCID}
}

\maketitle
%\vspace{-20pt}
\begin{abstract}
This paper investigates the use of unsupervised text-to-speech synthesis (TTS) as a data augmentation method to improve accented speech recognition. TTS systems are trained with a small amount of accented speech training data and their pseudo-labels rather than manual transcriptions, and hence unsupervised. This approach enables the use of accented speech data without manual transcriptions to perform data augmentation for accented speech recognition. Synthetic accented speech data, generated from text prompts by using the TTS systems, are then combined with available non-accented speech data to train automatic speech recognition (ASR) systems. ASR experiments are performed in a self-supervised learning framework using a Wav2vec2.0 model which was pre-trained on large amount of unsupervised accented speech data. The accented speech data for training the unsupervised TTS are read speech, selected from L2-ARCTIC and British Isles corpora, while spontaneous conversational speech from the Edinburgh international accents of English corpus are used as the evaluation data. Experimental results show that Wav2vec2.0 models which are fine-tuned to downstream ASR task with synthetic accented speech data, generated by the unsupervised TTS, yield up to 6.1\% relative word error rate reductions compared to a Wav2vec2.0 baseline which is fine-tuned with the non-accented speech data from Librispeech corpus.

\end{abstract}

\begin{IEEEkeywords}
Accented speech recognition, text-to-speech synthesis, data augmentation, self-supervised learning, Wav2vec2.0
\end{IEEEkeywords}

\vspace{-12pt}
\section{Introduction}
\label{sec:intro}
\vspace{-1pt}
Accented speech recognition is an important research topic of automatic speech recognition (ASR). Because of its importance, this research topic has been receiving attention and being addressed with various research approaches. In general, these approaches can be classified as accent-agnostic approaches, in which the modeling of accents inside the ASR systems is not made specific, and accent-aware approaches in which additional information about the accents of the input speech are used \cite{prabhu_emnlp_2023}. Among accent-agnostic approaches, adversarial learning was used to establish accent classifier and accent relabeling which led to performance improvement \cite{chen_icassp_2020, das_interspeech_2021, hu_icassp_2021}. In addition, similarity losses such as cosine losses or contrastive losses were used to build accent neutral models \cite{deng_interspeech_2021}. In accent-aware approaches, multi-domain training \cite{lucas_icassp_2023}, accent embeddings \cite{jain18_interspeech}, or accent information fusion \cite{wang_interspeech_2023} are among the approaches which have been investigated.

Text-to-speech synthesis (TTS) is an useful technology which can be used to improve ASR in a number of ways, for instance to improve the pre-training of self-supervised learning (SSL) models \cite{chen_asru_2021} or to improve the recognition of out-of-vocabulary words in end-to-end ASR \cite{zheng_icassp_2021}. TTS was also used as a data augmentation method to improve speech recognition for Librispeech task \cite{rosenberg_asru_2019} and low-resource speech recognition \cite{ueno_asru_2021, zhong_interspeech_2022, casanova_interspeech_2023}. More specifically, synthetic data were used for data augmentation in the context of low-resource ASR using conventional hybrid structure \cite{zhong_interspeech_2022} and to augment the training of RNN-T (recurrent neural network - transducer) ASR model \cite{fazel_interspeech_2021}. In \cite{casanova_interspeech_2023}, cross-lingual multi-speaker speech synthesis and cross-lingual voice conversion were applied to data augmentation for ASR. The authors showed that it is possible to achieve promising results for ASR model training with just a single speaker dataset in a target language, making it viable for low-resource scenarios \cite{casanova_interspeech_2023}. %The TTS model was trained using only one male speaker, but still produced strong results in zero-shot multi-speaker TTS and zero-shot voice conversion. 

While having been widely used in various ASR tasks, TTS, especially unsupervised TTS which is trained with unsupervised audio data \cite{ni_interspeech_2022}, has not been extensively studied as a data augmentation method in accented speech recognition. In a recent study on using TTS as data augmentation for accented speech recognition \cite{karakasidis_tsd_2023}, accented speech were generated by passing English text prompts through TTS system for a language corresponding to the target accent. For example, English text prompts passing through Spanish TTS will approximate Spanish-accented English. The study in \cite{karakasidis_tsd_2023} used commercial TTS systems whose training data were not accessible by users. 

In this paper, we investigate the use of unsupervised TTS as a data augmentation method to improve accented speech recognition. In our approach, we make use of a small amount of accented speech data which do not have manual transcriptions to train TTS systems. This approach enables the use of accented speech data without manual transcriptions to perform data augmentation for accented speech recognition. Indeed, from a small amount of unsupervised accented speech data used to train the TTS systems, we can generate larger amount of synthetic accented speech data once the TTS systems are trained. In this paper, from 58 hours of accented speech data, selected from two speech corpora of read speech: L2-ARCTIC \cite{zhao18b_interspeech} and British Isles \cite{demirsahin_lrec_2020}, we train unsupervised TTS and generate 250 hours more of synthetic accented speech data which help to achieve better gains on the evaluation data of spontaneous conversational speech from the Edinburgh international accents of English corpus (EdAcc) \cite{sanabria_icassp_2023}. 
%We show that the ASR models which are trained with synthetic accented speech data outperform a large Wav2vec2.0 baseline.

%These data are spoken in different speaking styles by speakers who differ from those in the evaluation data, but there is some overlaps on speakers' first languages in the two data. 
%In this paper, we investigate the use of unsupervised TTS as a data augmentation method to improve accented speech recognition. In our approach, we make use of a small amount of accented speech data to train TTS systems. These data are spoken in different speaking styles by speakers who differ from those in the evaluation data, but there is some overlaps on speakers' first languages in the two data. Such an approach facilitates the use of accented speech data without manual transcriptions to perform data augmentation for accented speech recognition. Indeed, from a small amount of accented speech data used to train the TTS systems, we can generate as much synthetic accented speech data as we need once the TTS systems are trained. In this paper, the accented speech training for training unsupervised TTS are read speech, selected from L2-ARCTIC \cite{zhao18b_interspeech} and British Isles \cite{demirsahin_lrec_2020} corpora, while spontaneous conversational speech from the Edinburgh international accents of English corpus (EdAcc) \cite{sanabria_icassp_2023} are used as evaluation data. We show that the ASR models which are trained with synthetic accented speech data outperform a large Wav2vec2.0 baseline.

The paper is organized as follows. In section \ref{sec:tts-based_data_augmentation}, details of the data augmentation for accented speech recognition based on unsupervised TTS are introduced. The training and inference of TTS systems are presented in section \ref{sec:tts}. Section \ref{sec:experiments} introduces the data used in the experiments, experimental results, and discussion. Finally, section \ref{sec:conclusion} concludes the paper.

%\vspace{-5pt}
%\section{TTS-based Data Augmentation for Accented Speech Recognition}
\section{Data Augmentation for Accented Speech Recognition based on Unsupervised TTS}
\label{sec:tts-based_data_augmentation}
%\vspace{-5pt}

\begin{figure}[t]	
\centering
		\includegraphics[width=0.95\columnwidth]{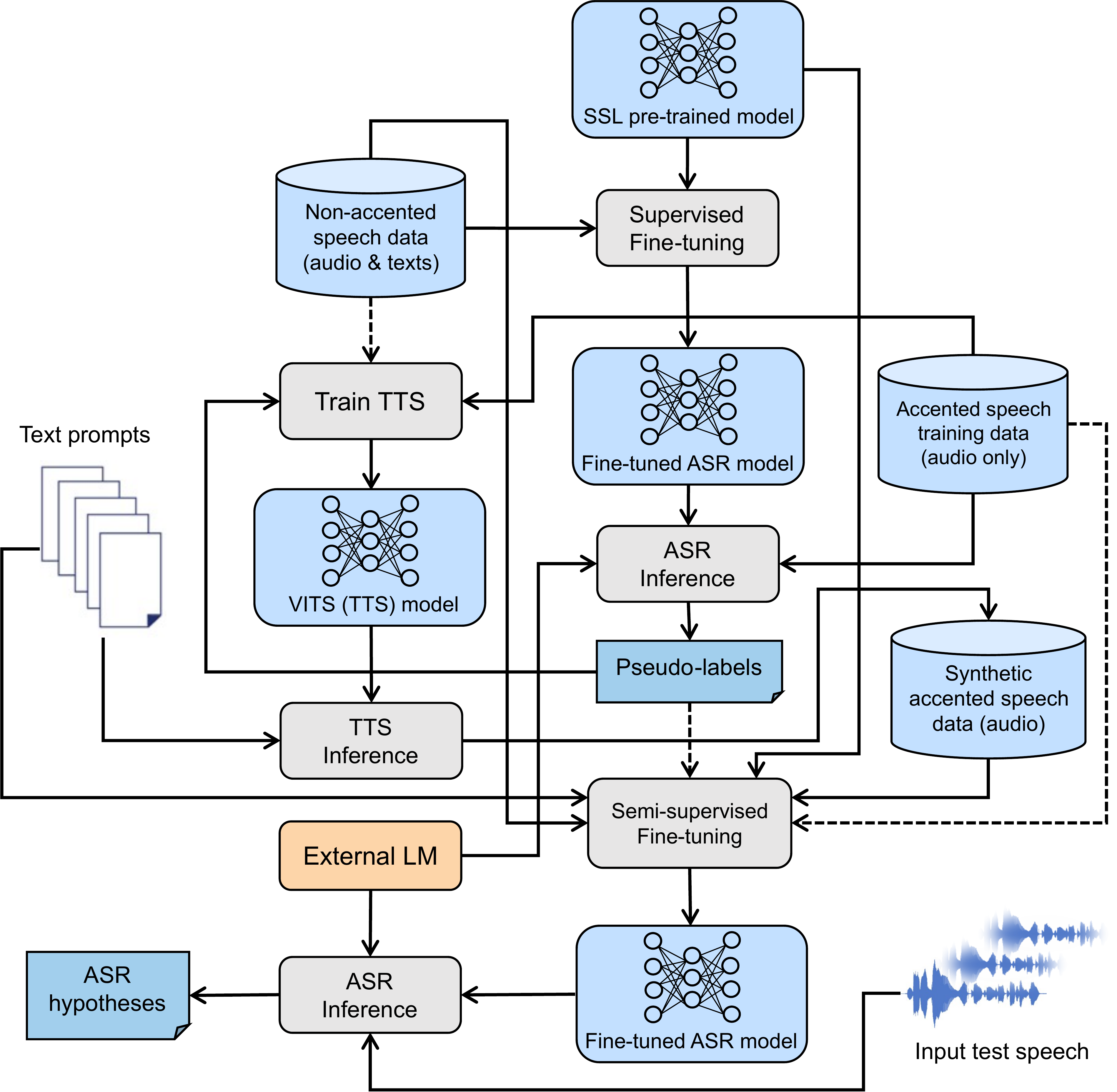}
	\caption{\label{fig:unsupervised_training} Unsupervised accented speech training data and theirs pseudo-labels are used to train unsupervised TTS. The pseudo-labels are generated by decoding the unsupervised accented speech training data using the baseline ASR model obtained from the supervised fine-tuning of the SSL pre-trained model with the supervised non-accented speech data. The unsupervised accented speech training data may be included in the semi-supervised fine-tuning for ASR, and the non-accented speech data may be used to train a TTS system.\vspace{-10pt}}
\end{figure}

\begin{figure}[t]
	\centering
		\includegraphics[width=0.95\columnwidth]{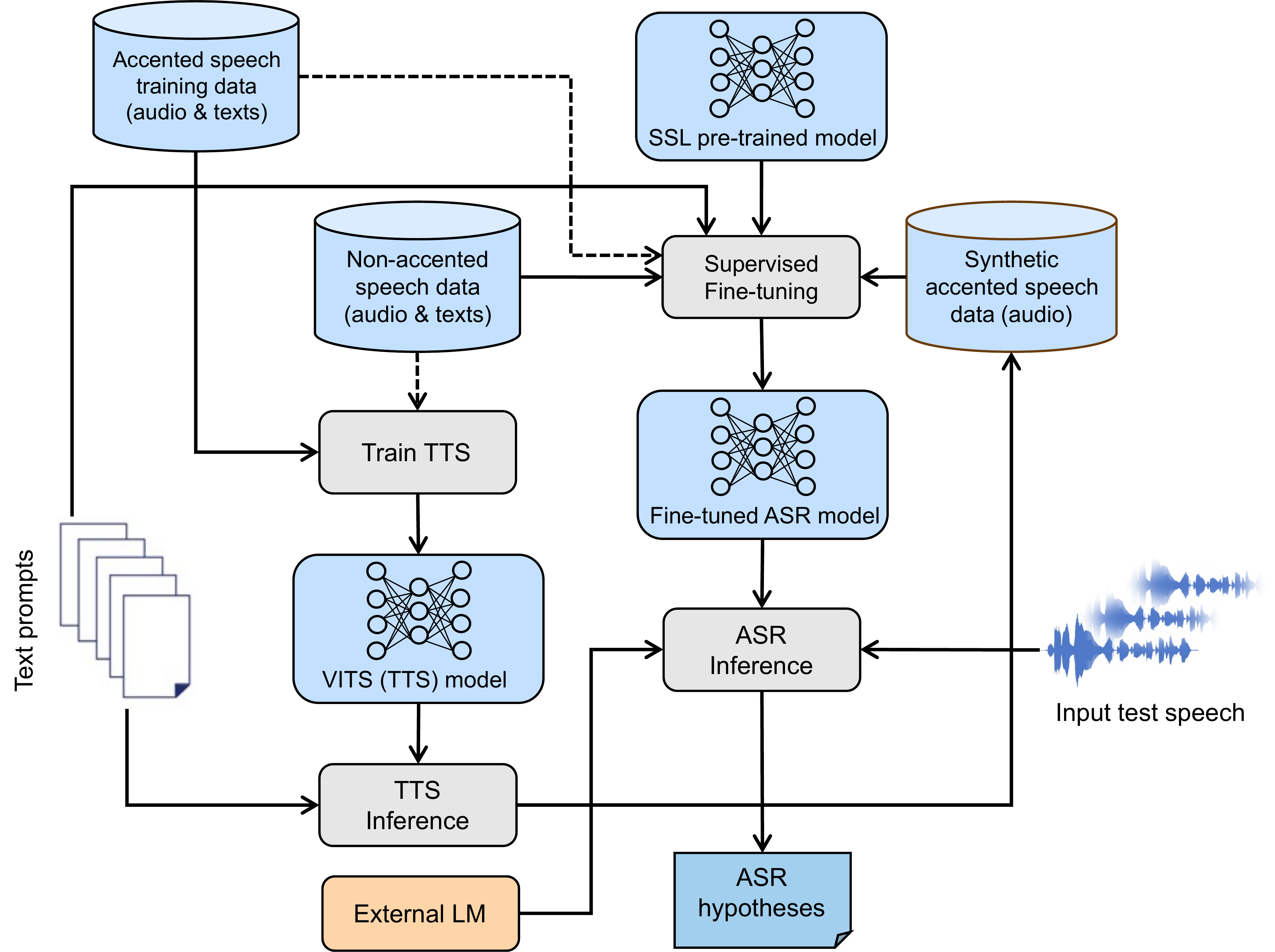}
	\caption{\label{fig:supervised_training} Supervised accented speech training data are used to train supervised TTS. These data may be included in the supervised fine-tuning for ASR, and the non-accented speech training data may be used to train a TTS system. The fine-tuned ASR model is used in the ASR inference, with an external language model (LM), to decode test speech. \vspace{-10pt}}
\end{figure}

We use Wav2vec2.0 SSL framework \cite{baevski_2020} for our experiments with accented speech recognition. Assume that a Wav2vec2.0 model was pre-trained via SSL on large amount of unsupervised speech data to cover various English accents and speakers, we can fine-tune this pre-trained model to downstream ASR task using available non-accented speech data. The non-accented speech data could be any available data which can be used to train ASR systems, for instance Librispeech training data \cite{panayotov_icassp_2015}. When using publicly available Wav2vec2.0 pre-trained models, we assume that only the models are available and their training data are not available.

%These non-accented speech training data can also be used to train TTS systems, which is a Variational Inference with adversarial learning for end-to-end Text-to-Speech (VITS) system \cite{kim2021conditional}, to generate synthetic speech data for data augmentation. Details about the VITS system used in this paper will be presented in section \ref{sec:tts}.

%These unsupervised accented speech data can primarily be used to train TTS systems, but can also be included in the training of ASR systems.

In addition to the non-accented speech data, we assume that a small amount of accented speech training data, named AccD, is available. These accented speech training data will be used to train the TTS systems. In accented speech recognition, it is not practical to find accented speech training data spoken in the same speaking styles and by the same speakers in the evaluation data. It is actually more viable to find accented speech training data which are spoken by speakers whose first languages are similar to those of the speakers in the evaluation data. Using these speech data to train TTS systems and generate more accented speech data for ASR training should create more accent variability, and hence, improve accented speech recognition performance.

%\vspace{-6pt}
\subsection{Unsupervised scenario}
\label{sec:unsupervised_scenario}
%\vspace{-2pt}

Fig. \ref{fig:unsupervised_training} shows unsupervised scenario where the manual transcriptions of the accented speech training data AccD are not available. Hence, pseudo-labels for the unsupervised accented speech training data are generated by decoding these data using the baseline ASR model obtained by fine-tuning the SSL pre-trained model with the supervised non-accented speech data. The unsupervised accented speech training data AccD and their pseudo-labels are then used to train TTS model, which is a Variational Inference with adversarial learning for end-to-end Text-to-Speech (VITS) model \cite{kim2021conditional}, to generate synthetic accented speech data for data augmentation. The unsupervised accented speech training data and their pseudo-labels may be included in the semi-supervised fine-tuning of the SSL pre-trained model. The ASR model obtained after the semi-supervised fine-tuning is used in the ASR inference to decode input test speech. The ASR inferences use an external language model (LM) when decoding audio data. 

A word-level 4-gram LM is used as external LM during ASR inferences. This 4-gram LM is trained on the manual transcriptions of Librispeech training data. The pre-training and fine-tuning of the Wav2vec2.0 models as well as the inference follow the same settings used for the LARGE Wav2vec2.0 models in \cite{baevski_2020}. These large models consist of 6 convolutional neural network (CNN) and 24 transformer layers, and have 350 millions parameters.

\subsection{Supervised scenario}
\label{sec:supervised_scenario}

For comparison, we also examine supervised scenario where the manual transcriptions of the accented speech training data AccD are available. In Fig. \ref{fig:supervised_training}, the accented speech training data AccD and their manual transcriptions can be directly used to train supervised TTS model. They may also be included in the training data which are used for the supervised fine-tuning of the SSL pre-trained model. Once the TTS model is trained, it can be used during the TTS inference to generate synthetic accented speech data using independent text prompts. Both the synthetic accented speech data and the text prompts can then be used for data augmentation in the supervised fine-tuning of the SSL pre-trained model.

\section{Text-to-speech Synthesis}
\label{sec:tts}

\begin{figure}[t]
	\centering
		\includegraphics[width=0.6\columnwidth]{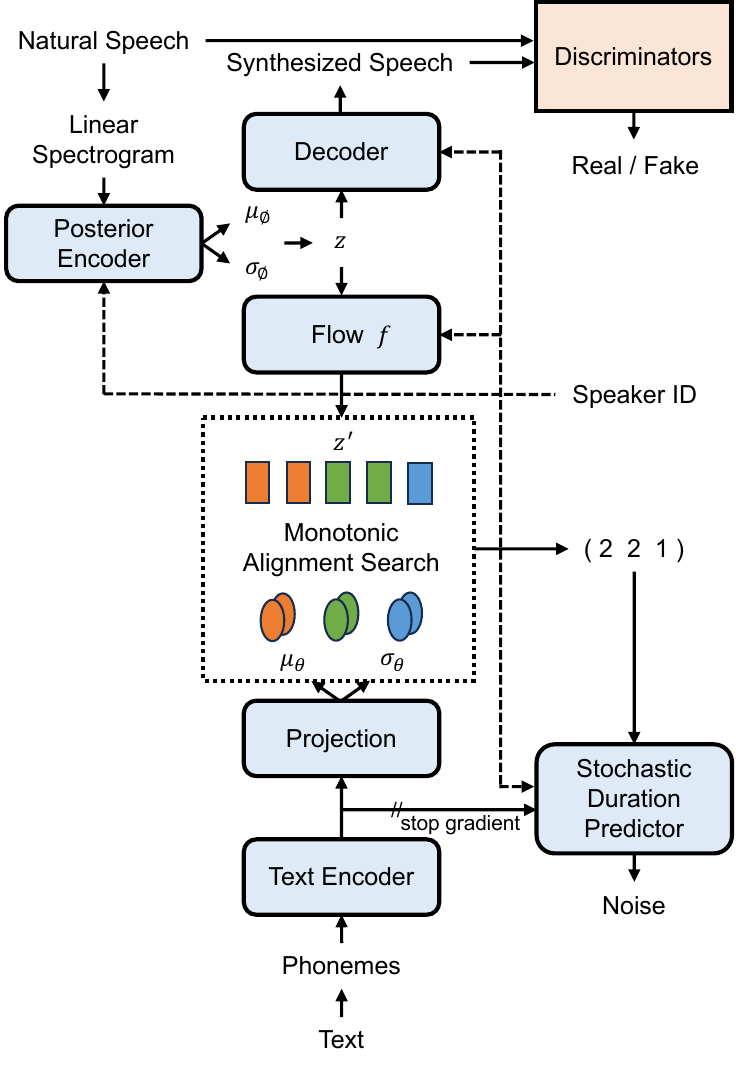}
	\caption{\label{fig:vits_training} Training of VITS parallel end-to-end TTS system. The input text can be either manual transcriptions or pseudo-labels.\vspace{-12pt}}
\end{figure}

VITS is an end-to-end multi-speaker TTS system which can generate high-quality waveforms \cite{kim2021conditional}. During the training of VITS (see Fig. \ref{fig:vits_training}), a Posterior Encoder encodes linear spectrogram from natural speech into a latent variable $z$ \cite{kingma2014autoencoding} which is then used in a Decoder to restore waveform. HiFi-GAN (Generative Adversarial Network) \cite{NIPS2014_5ca3e9b1}, a GAN-based neural vocoder \cite{kong2020hifi}, is used in the decoder to synthesize high-fidelity speech. The latent variable $z$ is also fed into the Flow $f$ which computes the Kullback-Leibler divergence with the Text Encoder outputs. The Flow $f$ is trained to remove speaker information and reduce posterior complexity \cite{rezende2015variational}. During training, speakers identities (IDs) are used to extract speaker embeddings for training multi-speaker TTS.  

\begin{figure}[t]
	\centering
		\includegraphics[width=0.43\columnwidth]{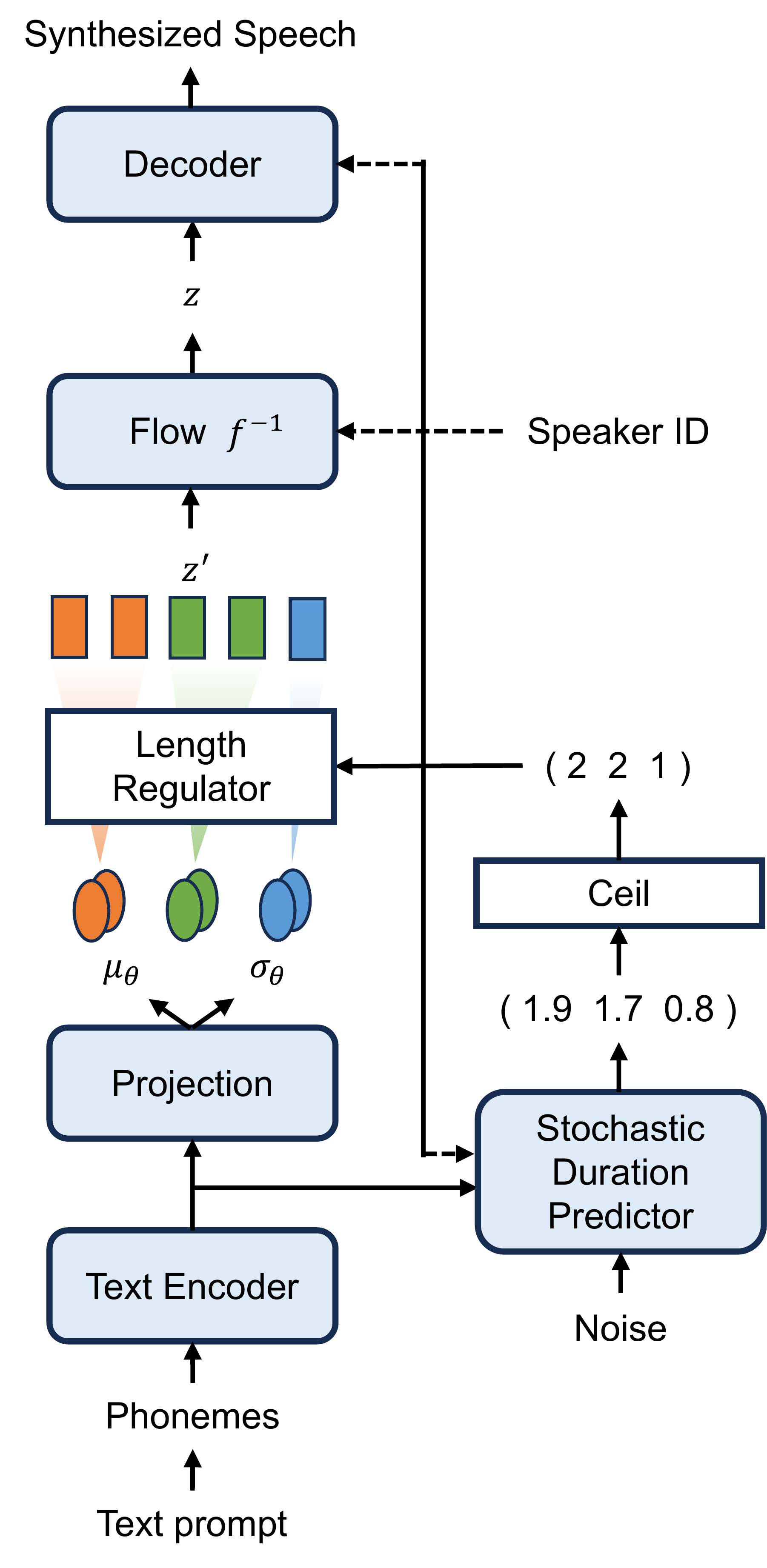}
	\caption{\label{fig:vits_inference} TTS inference using VITS model where text prompt and speaker ID are used as input. \vspace{-12pt}}
\end{figure}

During TTS inference (see Fig. \ref{fig:vits_inference}), an inverse transform $f^{-1}$ of the Flow $f$ is used to synthesize speech. The output of the Text Encoder is stretched by the Length Regulator based on the predicted duration, and then the sampled latent variable $z'\sim\mathcal{N}(z'; \mu_\theta(text), \sigma_\theta(text))$ is transformed by the inverse Flow $f^{-1}$ together with speaker information. Speech is subsequently generated by the decoder. All the VITS systems in this paper use the same architecture and are trained with the same number of training iterations, i.e. 300K. We observe that after 300K training iterations, the quality of the synthesized waveforms saturates. Greater details of the VITS  models and their implementation can be found in \cite{kim2021conditional}.

%\vspace{-5pt}
\section{Experiments}
\label{sec:experiments}

The Wav2vec2.0 model is pre-trained with more than 60K hours of unsupervised speech data from Libri-Light, CommonVoice, Switchboard, and Fisher corpora. These speech training data were spoken in various English accents. The non-accented speech data consist of 960 hours of training data from Librispeech corpus \cite{panayotov_icassp_2015} which include 2200 speakers. Although these speakers spoke US-English, we consider Librispeech as non-accented in the context of this study because US-English is only one of the English accents in the evaluation data. Subsequent experimental results confirm our assumption. %, and Librispeech is widely used in various ASR applications

\vspace{-5pt}
\subsection{Data}
\label{sec:data}

\subsubsection{Accented speech training data (AccD)}
\label{sec:accd_data}

\begin{table*}[t]
\begin{center}
\centering
\captionof{table}{Word error rates (WERs) on the development and test sets of the Edinburgh international accents of English corpus (EdAcc), and on the test-clean and test-other sets of Librispeech (LS) corpus.}
\label{tab:wers}
\begin{tabular}{|l|c|c|c|c|}
 \hline
\diagbox[width=9.25cm]{\small Fine-tuning data}{\small Test data} & \small \textrm{EdAcc} dev-set& \small \textrm{EdAcc} test-set & \small LS test-clean & \small LS test-other %\\ \hline
 \\\hline\hline
 \small LS 960h (M) (baseline in \cite{sanabria_icassp_2023}) & \small 33.4 & \small 36.1 & \small 2.9 & \small 5.6\\
\hline
 \small LS 960h (M) (our baseline) & \small 32.8 & \small 35.1 & \small 2.2 & \small 4.2\\
\hline\hline
\rowcolor[gray]{0.90} \small LS 960h (M) + AccD (P) & \small 32.4 & \small 34.6 & \small 2.1 & \small 4.1\\
\hline
\rowcolor[gray]{0.90} \small LS 960h (M) + AccD (M) & \small 31.1 & \small 33.4 & \small 2.1 & \small 4.0\\
\hline\hline
 \small LS 960h (M) + TTS-LS 960h (M) & \small 31.4 & \small 33.8 & \small 2.1 & \small 4.0\\
\hline
\small LS 960h (M) + TTS-AccD (P) & \small 31.0 & \small 33.2 & \small 2.1 & \small 4.1\\
\hline
\small LS 960h (M) + TTS-AccD (M) & \small 30.8 & \small 33.0 & \small 2.1 & \small 4.1\\
\hline\hline
\rowcolor[gray]{0.90} \small LS 960h (M) + AccD (P) + TTS-AccD (P) + TTS-LS 960h (M) & \small 30.8 & \small 33.2 & \small 2.1 & \small 4.2\\
\hline
\rowcolor[gray]{0.90} \small LS 960h (M) + AccD (M) + TTS-AccD (M) + TTS-LS 960h (M) & \small 30.4 & \small 32.7 & \small 2.1 & \small 4.1\\
\hline
\end{tabular}
\end{center}
\vspace{-12pt}
\end{table*}

We combine data from the L2-ARCTIC corpus \cite{zhao18b_interspeech} and the British Isles corpus \cite{demirsahin_lrec_2020} as accented speech training data. These are corpora of read speech which were recorded in controlled environments. The L2-ARCTIC corpus is a speech corpus of non-native English which contains 26,867 utterances from 24 non-native English speakers with equally distributed number of speakers per accent. The total duration of the corpus is 27.1 hours, with an average of 67.7 minutes of speech per speaker. On average, each utterance is 3.6 seconds in duration. The utterances in L2-ARCTIC are spoken in 6 non-native accents: Arabic, Chinese, Hindi, Korean, Spanish, and Vietnamese.

The British Isles corpus includes speech utterances recorded by volunteers speaking with different accents of the British Isles, namely Ireland, Scotland, Wales, the Midlands, Northern, and Southern of England. The corpus consists of 17,877 utterances spoken by 120 speakers of which 49 are female and 71 are male. The total duration of the corpus is 31 hours. When being decoded in the unsupervised scenario, the WERs of the pseudo-labels obtained on the L2-ARCTIC and British Isles training data are 10.7\% and 10.2\%, respectively. 

%\vspace{-1pt}
\subsubsection{Evaluation data}
\label{sec:evaluation_data}

We use the development and test sets from the Edinburgh international accents of English corpus (EdAcc) \cite{sanabria_icassp_2023}, which consist of spontaneous conversational speech, as evaluation data. The corpus includes a wide range of first- and second-language varieties of English in the form of dyadic video call conversations between friends. The conversations range in durations from 20 to 60 minutes. These conversations are segmented into shorter utterances based on manual annotations and are then separated into development and test sets which consist of 9079 and 8494 utterances, respectively. In total, the development set contains 14 hours and the test set contains 15 hours of speech. There are more than 40 self-reported English accents from 51 different first languages. The statistics and analyses show that EdAcc is linguistically diverse and challenging for current English ASR systems \cite{sanabria_icassp_2023}. With more than 40 English accents, the EdAcc corpus covers English accents from four continents, including Africa, America, Asia, and Europe. The conversations were manually transcribed by professional transcribers to obtain manual transcriptions which are used in the evaluation.

%\vspace{-1pt}
\subsubsection{Synthetic speech data}
\label{sec:synthetic_data}

Synthetic speech data are generated using the TTS systems and English text prompts. The text prompts used in the TTS inference are selected from the manual transcriptions of the training data in three speech corpora: LJSpeech \cite{ljspeech17}, TED-LIUM \cite{rousseau2012tedlium}, and VCTK \cite{vctk19}. The objective of selecting text prompts from independent TTS and ASR corpora is to ensure that these prompts are not related to the evaluation data and are phonetically balanced, since they were designed for TTS and ASR applications. In total, there are 120K text prompts resulting in 250 hours of synthetic speech data which are spoken by the speakers presented in the training data of the TTS systems.% The text prompts are used as manual transcriptions of the synthetic audio data used in the data augmentation for the fine-tunings.

\vspace{-5pt}
\subsection{Results \& Discussion}
\label{sec:results}

Experimental results, in terms of WERs, are shown in Table \ref{tab:wers}. In Table \ref{tab:wers}, the WERs computed on the EdAcc development \& test sets and the Librispeech (LS) test-clean \& test-other sets are shown. The ASR models in Table \ref{tab:wers} are fine-tuned from one Wav2vec2.0 pre-trained model, which was pre-trained on the unsupervised training data of Libri-Light, Common Voice, Switchboard, and Fisher, using different fine-tuning data. The abbreviations used in Table \ref{tab:wers} have the meaning as follows:
\begin{itemize}
\justifying
 \item LS 960h (M): 960 hours of training speech from Librispeech, manual (M) transcriptions are used as labels.
	\item AccD (P), AccD (M): 58 hours of accented speech training data, using either pseudo-labels (P) or manual (M) transcriptions as labels.
	\item TTS-LS 960h (M): 250 hours of synthetic non-accented speech data generated by TTS system trained on LS 960h (M) data. The speakers are from the LS 960h data.
	\item TTS-AccD (P), TTS-AccD (M): 250 hours of synthetic accented speech data generated by TTS systems trained on either AccD (P) or AccD (M) data, with speakers from the AccD data.
\end{itemize}

We build a baseline model by fine-tuning the Wav2vec2.0 pre-trained model with the LS 960h (M) data. The Wav2vec2.0 pre-trained model and the fine-tuning data that we use are the same as those used to train the baseline model in \cite{sanabria_icassp_2023}. We will compare the results with our baseline model which has lower WERs, compared to those of the Wav2vec2.0 model reported in \cite{sanabria_icassp_2023}, on the development and test data of both EdAcc and Librispeech (see Table \ref{tab:wers}). Combining the unsupervised accented speech training data AccD (P) with the non-accented speech data LS 960h (M) to fine-tune the pre-trained model yields 1.2\% and 1.4\% relative WER reductions on EdAcc dev and test sets, respectively, while the respective relative WER reductions on these sets are 5.2\% and 4.8\% when the supervised accented speech training data AccD (M) are used. %

When the synthetic non-accented speech data TTS-LS 960h (M) which are generated by the supervised TTS system, trained on the non-accented speech data LS 960h (M) with manual transcriptions, are included in the fine-tuning, 4.3\% and 3.7\% relative WER reductions are obtained on the EdAcc dev and test sets, respectively. Since the synthetic non-accented speech data TTS-LS 960h (M) are spoken by the same speakers in the LS 960h (M) data, the relative WER reductions are made mainly thanks to more acoustic realizations, based on the independent text prompts, are added to the fine-tuning data from the synthetic non-accented speech data. Larger gains are obtained when the synthetic accented speech data TTS-AccD (P) and TTS-AccD (M) are used, even though the amount of data and the number of speakers in the AccD data used to train TTS systems are much smaller compared to those of the non-accented speech data LS 960h (M): 58 hours compared to 960 hours, and 144 speakers compared to 2200 speakers. More specifically, the TTS-AccD (P) data generated by unsupervised TTS help to achieve 5.5\% and 5.4\% relative WER reductions on the EdAcc dev and test sets, respectively, while  the TTS-AccD (M) data generated by supervised TTS help to achieve 6.1\% and 6.0\% relative WER reductions on the EdAcc dev and test sets, respectively.

When the accented speech training data AccD and all the synthetic speech data are combined with the non-accented speech data to fine-tune the pre-trained model, further gains are obtained. In the unsupervised scenarios, 6.1\% and 5.4\% relative WER reductions are obtained on the EdAcc dev and test sets, respectively, while the respective relative WER reductions obtained on these sets in the supervised scenario are 7.3\% and 6.8\%, respectively. Actually, using natural accented speech training data and synthetic accented speech data improves the performance on EdAcc dev and test sets but does not harm or improve the ASR performance on Librispeech test sets. This confirms that considering Librispeech training data as non-accented speech data in our experiments is relevant.

\vspace{-2pt}
\section{Conclusion}
\label{sec:conclusion}

\vspace{-2pt}
Unsupervised TTS, trained on unsupervised accented speech training data, was used to generate synthetic accented speech data for data augmentation in accented speech recognition. Experiments showed that the Wav2vec2.0 models which used the synthetic accented speech data yielded up to 6.1\% relative WER reductions compared to a large Wav2vec2.0 baseline. These gains are close to those obtained in the supervised scenario. The results demonstrate that unsupervised accented speech data, even when available in limited quantities and are spoken in different styles by speakers who differ from those in the evaluation data, can be effectively used to train TTS systems for data augmentation. This approach improves accented speech recognition, particularly when the speakers in the unsupervised accented speech data and those in the evaluation data have some overlaps on speakers’ first languages.
%Synthetic accented speech data were generated by unsupervised TTS and were then combined with the accent-agnostic training data to train ASR systems. 

\vspace{-2pt}
\bibliographystyle{IEEEbib}
\bibliography{refs}

\end{document}